\documentclass[letterpaper, 10 pt, conference]{ieeeconf}  

\IEEEoverridecommandlockouts                              

\usepackage{cite}
\usepackage{amsmath,amssymb,amsfonts}
\usepackage{textcomp}
\usepackage{multicol}
\usepackage{lipsum}
\usepackage{longtable}
\usepackage{vcell}
\usepackage{wrapfig}
\usepackage{epsfig}
\usepackage{graphicx}
\usepackage{subfigure} 
\usepackage{booktabs}
\usepackage{multirow}
\usepackage{hhline}
\usepackage{textcomp}
\usepackage{url}
 \usepackage[leftcaption]{sidecap}

\usepackage{xcolor, soul}
\usepackage{colortbl}
\usepackage{subcaption}
\usepackage{etoolbox}

\usepackage{array}
\usepackage{ulem}
\usepackage[linesnumbered,ruled,vlined]{algorithm2e}
\usepackage{caption}
\usepackage{booktabs}

\usepackage{setspace}
\title{\LARGE \bf
SkyVLN: Vision-and-Language Navigation and NMPC Control for UAVs in Urban Environments
}

\author{Tianshun Li$^{1}$, Tianyi Huai$^{1}$, Zhen Li$^{1}$, Yichun Gao$^{1}$, Haoang Li$^{2}$ and Xinhu Zheng$^{3, \ast}$ 
\thanks{$^\ast$ Corresponding author.}
\thanks{$^{1}$ Tianshun Li, Tianyi Huai, Zhen Li, and Yichun Gao are with Intelligent Transportation, Hub of Systems,
        The Hong Kong University of Science and Technology (Guangzhou), China
        {\tt\small \{tli449, thuai231, zli619, ygao514\}@connect.hkust-gz.edu.cn}}%
        \thanks{$^{2}$Haoang Li is with the Intelligent Transportation Thrust and Robotics and Autonomous Systems Thrust, Systems Hub, The Hong Kong University of Science and Technology (Guangzhou), China
        {\tt\small haoangli@hkust-gz.edu.cn}}
        \thanks{$^{3}$Xinhu Zheng is with the Intelligent Transportation Thrust, Systems Hub, Internet of Things Thrust, Information Hub, The Hong Kong University of Science and Technology (Guangzhou), China
        {\tt\small xinhuzheng@hkust-gz.edu.cn}}%
}

\begin{document}

\setlength{\abovecaptionskip}{0pt} 
\setlength{\belowcaptionskip}{2pt} 
\setstretch{0.9}

\maketitle
\thispagestyle{empty}
\pagestyle{empty}

\begin{abstract}
Unmanned Aerial Vehicles (UAVs) have emerged as versatile tools across various sectors, driven by their mobility and adaptability. This paper introduces SkyVLN, a novel framework integrating vision-and-language navigation (VLN) with Nonlinear Model Predictive Control (NMPC) to enhance UAV autonomy in complex urban environments. Unlike traditional navigation methods, SkyVLN leverages Large Language Models (LLMs) to interpret natural language instructions and visual observations, enabling UAVs to navigate through dynamic 3D spaces with improved accuracy and robustness.
We present a multimodal navigation agent equipped with a fine-grained spatial verbalizer and a history path memory mechanism. These components allow the UAV to disambiguate spatial contexts, handle ambiguous instructions, and backtrack when necessary. The framework also incorporates an NMPC module for dynamic obstacle avoidance, ensuring precise trajectory tracking and collision prevention.
To validate our approach, we developed a high-fidelity 3D urban simulation environment using AirSim, featuring realistic imagery and dynamic urban elements. Extensive experiments demonstrate that SkyVLN significantly improves navigation success rates and efficiency, particularly in new and unseen environments. 
\end{abstract}

\section{INTRODUCTION}
\label{sec:introduction}

Unmanned Aerial Vehicles have been a focus of attention recently due to their remarkable autonomy, mobility, and adaptability, enhancing a wide array of applications, including surveillance \cite{li2021networked}, monitoring, search and rescue, healthcare, and wireless network provisioning \cite{tian2025uavsmeetllmsoverviews}. Concurrently, UAVs are enhancing logistics by optimizing route planning and inventory management. This not only streamlines warehouse operations but also significantly boosts delivery efficiency 
\cite{al2020comprehensive}.

Among the recent technological advancements, Large Language Models have emerged as a focal point of interest due to their ability to learn from application behaviors and refine existing systems \cite{rasley2020deepspeed}. Their adaptive learning features allow for ongoing refinement of operational strategies based on real-time data inputs, thereby significantly improving decision-making processes \cite{kurunathan2023machine}. LLMs are capable of processing complex multi-modal inputs, including visual information and natural language instructions \cite{Zhou2023NavGPTER, Zhou2024NavGPT2UN}. They achieve task planning and common-sense reasoning through pre-trained universal representations of vision and language. 
Inspired by the potential of integrating LLM with navigation \cite{park2023visual}, it is compelling that we can explore how LLMs can be integrated into UAVs' visual language navigation tasks to enhance not only the navigation performance but also the explainability and robustness.

\begin{figure}[ht]
    \centering
    \includegraphics[width=0.97\linewidth]{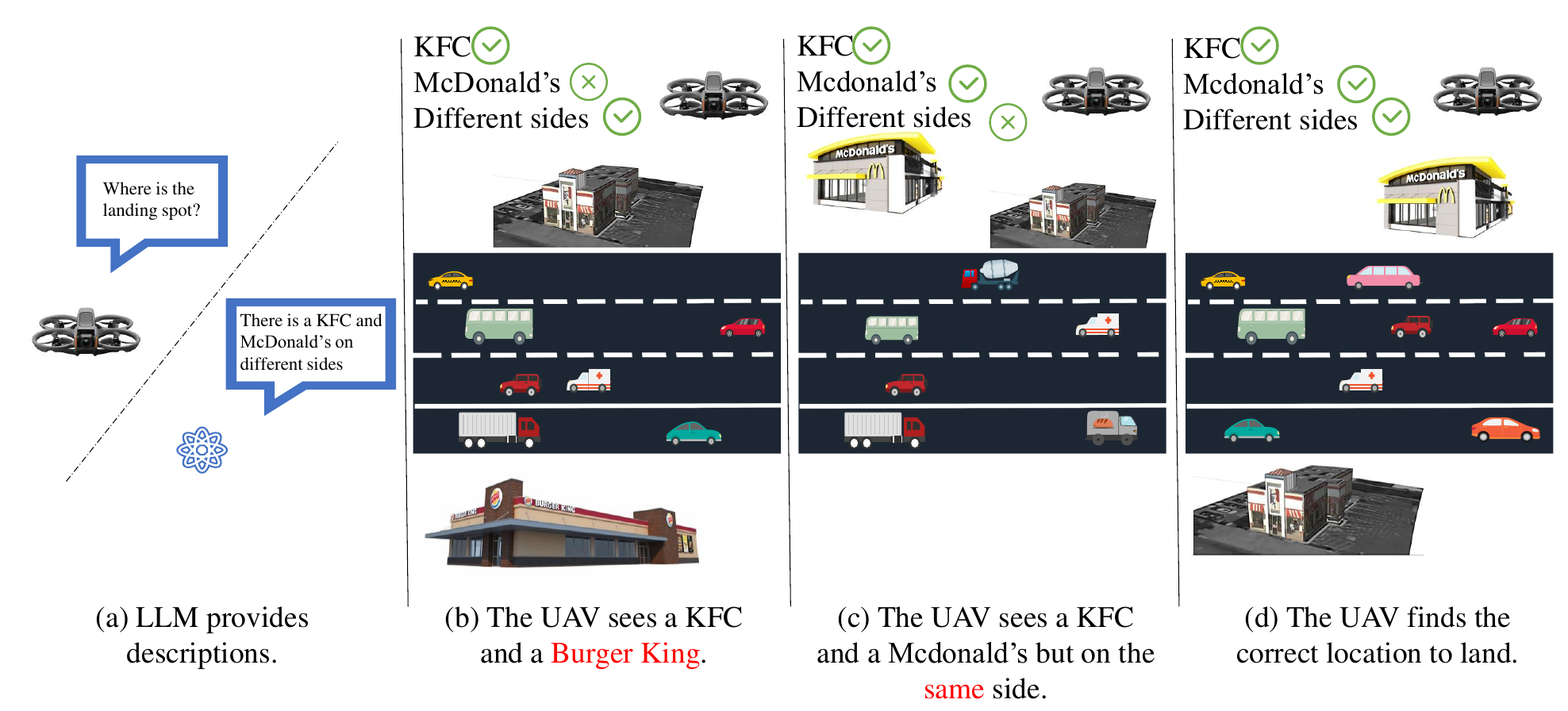}
    \caption{Vision and language navigation for UAVs with place recognition. LLM gives UAV verbal descriptions of designed landing spot (a). The UAV compares its visual observations with LLM’s descriptions (b)-(d) and reasons about their accuracy, confirming (d) as the correct place.}
    \label{fig:VLN Task}
    \vspace{-0.8em}
\end{figure}

Specifically, humans typically synthesize natural language descriptions with visual landmarks to clarify their spatial context before navigating. Similarly, a UAV performing VLN tasks mimics this cognitive paradigm, as illustrated in Figure. \ref{fig:VLN Task}. For example, the UAV agent may receive a description like, ``There is a KFC and McDonald’s on different sides," which it uses to identify a landing spot. However, this description lacks explicit navigation commands and merely lists surrounding landmarks without providing spatial context. The UAV agent must then compare its visual observations with the linguistic cues from the LLM through spatial-semantic consistency validation. A mismatch in the spatial relationship of the landmarks (e.g., the positions of KFC and McDonald’s) indicates that the UAV is in the wrong place, while a match confirms it is in the correct location.  

Although combining linguistic cues with visual place recognition offers an intuitive approach to navigate, several fundamental challenges remain unexplored: (a) how LLMs or VLMs can enhance cross-modal grounding, and (b) the quantitative efficacy of introducing the visual location with the language-derived spatial descriptions.

When doing a visual comparison, humans tend to gather contextual information across the observations iteratively\cite{LyuMultimodalLLMsMeetPlaceRecognition2024}. However, existing LLMs combined with VPR often fail to capture this iterative process, relying instead on a more static, one-step comparison. This limitation suggests that current methods may not fully leverage the dynamic, context-rich reasoning that humans employ. To address this, we propose a comparing-then-reasoning paradigm for human-like visual-language alignment. Multimodal LLMs describe the delta of each image pair and then use all the textual descriptions for the final reasoning stage to determine the best candidate. This approach aims to more closely mimic human visual comparison strategies, potentially improving the accuracy and robustness of visual location.

However, without task-specific fine-tuning, it is challenging for multimodal LLMs to distinguish landmark-relevant details and landmark-irrelevant details, for instance, the building patterns vs. shadows cast by clouds, while humans can easily extract landmark-critical features for landmark-relevant details. 
Multimodal LLMs tend to generate non-factual descriptions due to their tendency to hallucinate details that are not present in the input data. When dealing with complex contexts, such as a description with multiple candidate locations, LLMs often struggle to keep focus on the most relevant landmark information, leading to potential oversight of landmark-relevant details. Injecting human knowledge into multimodal LLMs via specific prompts is proven to be effective and performs well in outdoor navigation tasks\cite{chen2019touchdown}, yet its environments are static.

\begin{figure}
    \centering
    \includegraphics[width=0.98\linewidth]{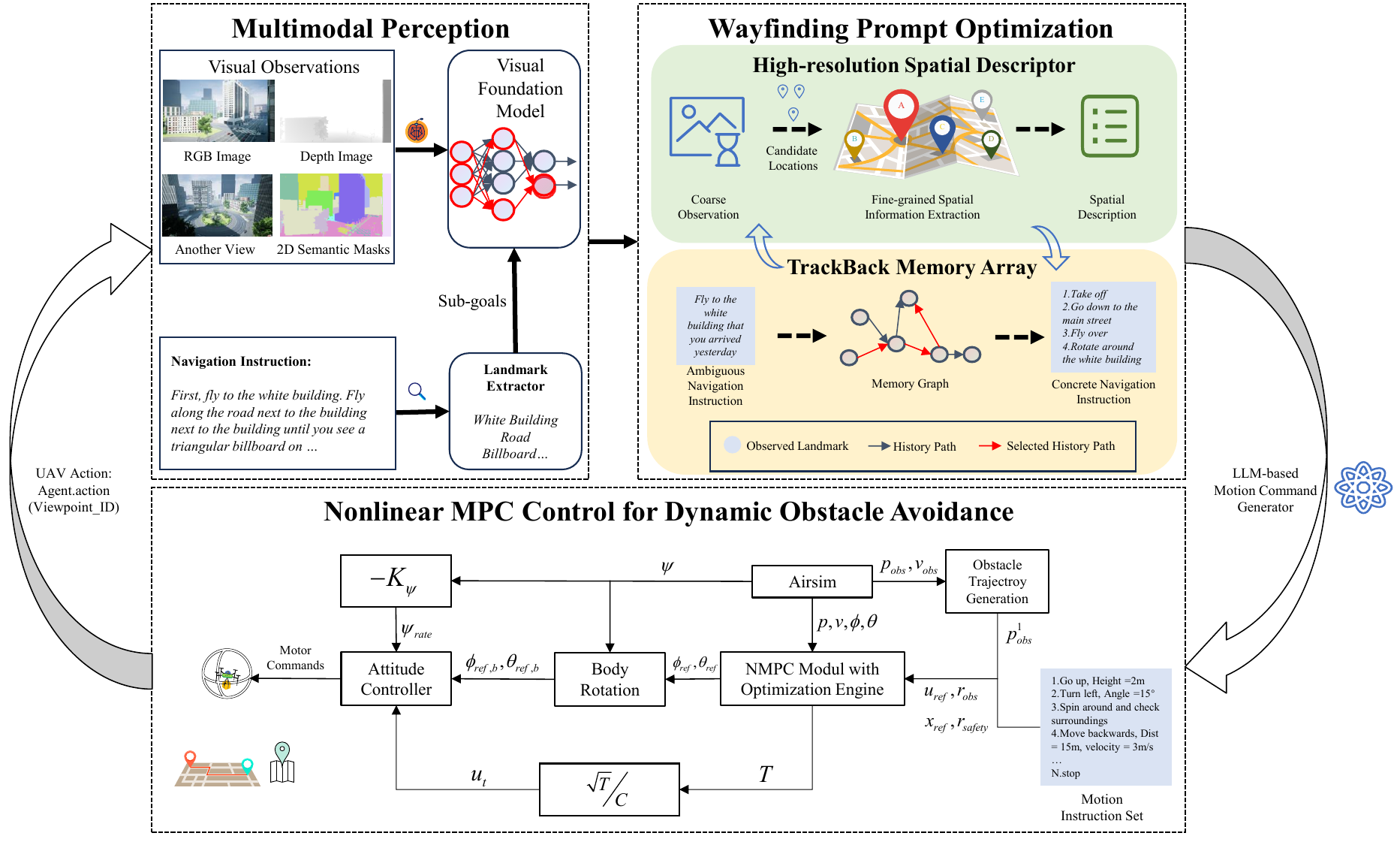}
    \caption{The overall architecture of the VLN agent, which can take both the instruction and visual perceptions into consideration. This solution does not need supervised training. WPO is proposed to further refine the localization precision.}
    \label{fig:VLN agent}
    \vspace{-2em}
\end{figure}

Navigation in the sky is significantly different from that on the ground in several aspects:

\begin{itemize}
    \item Expanded action space, from 2D to 3D. Aerial navigation needs to be conducted in three-dimensional space, where drones have to deal with more complex altitude and spatial relationships. Compared to conventional ground VLN \cite{KrantzBeyondtheNavGraph2020,ku2020roomacrossroommultilingualvisionandlanguagenavigation}, UAVs need to additionally take actions such as “rise up” and “pan down”. Moreover, multirotors can move left/right without turning its head, and pitch, roll, yaw.
    \item Complex visual localization. In urban environments, dense high-rise buildings and severe electromagnetic interference can obstruct traditional GNSS signals or cause multipath effects, significantly degrading navigation accuracy. Therefore, visual localization methods are essential not only to supplement depth and pose information but also to provide accurate height data.
    \item Dynamic obstacle avoidance. Aerial navigation involves longer flight paths, and long-distance flights require avoiding dynamic obstacles. Weather and lighting conditions also make aerial navigation more challenging.
\end{itemize}

To release humans from manually operating aerial vehicles and to fill the research gap in the field of navigation in the sky, we propose a city-level UAV-based vision-and-language navigation task, named SkyVLN.
The contributions of this paper are summarized as follows:
\begin{itemize}
    \item SkyVLN is a novel vision-and-language navigation framework that combines vision-language models with prompt optimization and nonlinear MPC to enhance UAV operations in complex urban environments. This integration allows UAVs to interpret natural language instructions and visual perceptions for both improved navigation performance and decision explainability. Through extensive evaluations, the SkyVLN framework demonstrates significant improvements in navigation success rates and efficiency, particularly in new environments. 
    
    \item In the design of nonlinear MPC strategy, a variety of constraints are considered, including velocity, attitude limitations, angular velocity of the attitude, and collision avoidance requirements. By predicting future system behavior and optimizing control inputs, NMPC is able to achieve the desired trajectory tracking while satisfying the physical constraints.

    \item The spatial verbalizer provides detailed descriptions of landmarks and their spatial relationships, while the history path memory helps the UAV maintain context over time and backtrack when necessary. This combination significantly enhances the UAVs' ability to handle ambiguous instructions and complex spatial reasoning tasks, leading to better navigation performance in unseen environments.
\end{itemize}

The rest of the paper is organized as follows. Section \ref{sec:Related} reviews existing literature on UAV navigation, vision-and-language tasks, and the integration of language models in robotic systems. Section \ref{sec:Task} presents our proposed framework, SkyVLN, detailing the design of the embodied navigation agent, the fine-grained spatial verbalizer, and the history path memory mechanism. We also explain the macro-action description approach used for interpreting navigation instructions. Section \ref{sec:Platform} focuses on the development of our high-fidelity 3D simulator, emphasizing its capabilities for realistic imagery and dynamic urban environment configurations. Section \ref{sec:Experiment} presents the experimental setup, evaluation metrics, and results of our navigation experiments. We analyze the performance of the SkyVLN framework in various scenarios and compare it with baseline methods. Finally, section \ref{sec:conclusion} concludes with a summary of our contributions.

\section{Related Work}
\label{sec:Related}

We categorize three types of closely related work in visual language navigation for UAVs: direct navigation, instruction-level navigation, and end-to-end navigation.
\subsubsection{Direct}
Traditional UAV navigation often employs direct control methods, where pre-programmed commands are sent to the UAV to control its movements \cite{Hashim_2025}. These methods are typically based on pre-defined paths or real-time sensor data and are crucial for tasks requiring precise maneuvering \cite{10294277}. Direct control methods are efficient for tasks with well-defined objectives and environments. But they lack flexibility in urban areas with collapsed structures or complex electromagnetic conditions or unexpected disturbances, which can lead to unstable flight, limiting their capacity for dynamic interaction. Direct control lacks the autonomy to handle tasks independently, making it less suitable for long-duration or remote operations.

\subsubsection{Instruction Level}
Instruction-level control involves interpreting high-level commands or instructions to guide UAV behavior \cite{wang2024visionbaseddeepreinforcementlearning}. This approach allows for more human-like interaction with the UAV, enabling it to adapt to changing environments based on natural language instructions \cite{choutri2022multi}. LLMs, such as those employing transformer architectures, have been instrumental in enhancing instruction-level control by providing the capability to understand and generate human-like text \cite{tagliabue2023real}, which simplifies UAV control and allows handling complex, real-time mission adjustments \cite{wang2024realisticuavvisionlanguagenavigation}. Building upon BERT's capabilities \cite{devlin2018bert}, a novel framework known as the Language Model-based fine-grained Address Resolution (LMAR) was introduced in \cite{luo2024language}. Yet, natural language instructions can be ambiguous, leading to misinterpretation by the UAV, which can result in incorrect actions or failures. 

\subsubsection{End-to-End}
End-to-end UAV navigation systems leverage machine learning techniques to map sensory inputs directly to control actions, eliminating the need for handcrafted feature extraction and decision-making processes \cite{10817801}. These systems have shown promise in handling complex data structures and providing better generalization capabilities across different environments \cite{GYAGENDA2022104069}. End-to-end policies require large amounts of high-quality training data to achieve good performance, which can be difficult to obtain, especially for specific or niche applications. These models may not generalize well to new or unseen environments, leading to performance degradation.

Unlike conventional control systems that are heavily dependent on specific environment and operating skills and limited by the accuracy of instruction data provided by users, SkyVLN is proposed to leverage robust visual features from off-the-shelf vision foundation models to obtain candidate locations and uses LLMs for language-based reasoning and decision-making. A Nonlinear Model Predictive Control strategy is utilized for UAVs to navigate complex environments efficiently, ensuring stability and precision in trajectory tracking.

\section {The SkyVLN Task} \label{sec:Task}
Our system architecture has been presented in Figure. \ref{fig:VLN agent}. Our system comprises a multimodal perception module and an action decision module based on nonlinear model predictive control. The agent first perceives the environment through visual and lingual information. Then, the Wayfinding Prompt Optimization (WPO) extracts richer spatial information from the perception results. Besides, WPO leverages the memory layer storing historical trajectories to provide more navigation clues for ambiguous navigation tasks. Finally, the LLM motion generator outputs its current thoughts and actions by using the navigation prompts from the aforementioned two modules along with the system and action prompt as inputs.

Formally, at the beginning of each episode, the agent is placed in an initial pose \(P=[x,y,z,u,v,w,\phi ,\theta ,\psi ,p,q,r]^{T}\), where $ [x,y,z]^{T}$ denotes the agent’s position and $ [\phi ,\theta ,\psi]^{T}$ represent the roll, pitch, yaw portion of the agent’s orientation, $[u,v,w]^{T}$ and $[p,q,r]^{T}$ are their corresponding derivatives, respectively. Then given a natural language instruction $X =< {{w}_{1}},{{w}_{2}},...,{{w}_{L}} >$, where \(L\) is the length of instruction and \({w}_{i}\) is a single word token, the agent is required to predict a series of actions. The agent can take both the instruction and visual perceptions into consideration. Although our adopted simulator can provide panoramic observations, here we follow the most robotic navigation tasks \cite{JacobBeyond2020} setting to limit our baseline agent to the access of its front view perceptions (RGB, depth and semantic images) ${{V}_{t}}=\left\{ v_{t}^{R},v_{t}^{D},v_{t}^{S} \right\}$. The agent needs to rotate to obtain other views. Navigation ends when the agent predicts a \textit{STOP} action or reaches a pre-defined maximum action number.

\subsection{Multimodal Perception}
The drone is equipped with a wide-angle camera at the front of its body, which can rotate up and down 90 degrees along the pitch axis but cannot rotate left or right. To thoroughly perceive its surroundings, the drone needs to capture a panoramic image by rotating itself. 
An off-the-shelf vision-language model (VLM) namely GroundingDINO \cite{oquab2024dinov2learningrobustvisual} is then employed to detect landmarks in each image coarsely. If the same landmark appears in multiple views, the perspective in which the landmark has the highest score is designated as the landmark’s observation viewpoint.

In the realm of navigational guidance, the instruction \(T\) contains several key phrases that indicate the need for directional changes. The initial phase of our process involves identifying these pivotal phrases. To achieve this, we employ a pre-trained LLM, renowned for its exceptional ability to understand context and excel in zero-shot learning scenarios, as our tool for text analysis. This sophisticated model will meticulously identify and extract the landmark phrases from the instruction, compiling them into a set \(L\), which consists of elements \(l_1\) through \(l_n\). The sub-goal extraction process is represented as:
\begin{equation}
L=\operatorname{LLM}\text { } (T,\text { prompt})
\end{equation}

\begin{figure}[ht]
    \centering
    \includegraphics[width=1\linewidth]{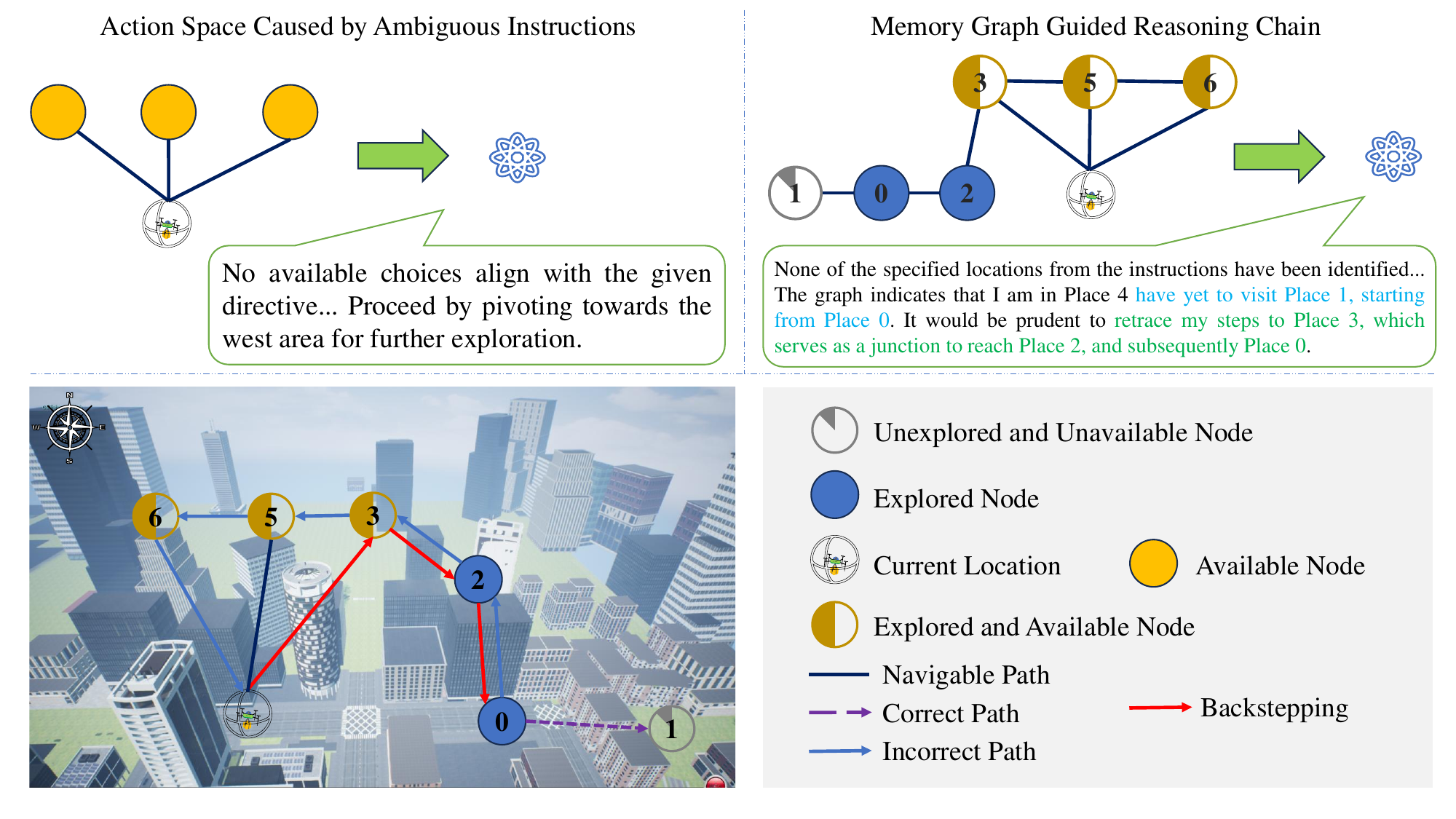}
    \caption{A comparison of the thinking process of the GPT agent without and with memory graph. Given an ambiguous description with ambiguous instructions, the agent may explore aimlessly, especially when navigation errors have already occurred.}
    \label{fig:Ambiguous}
    \vspace{-0.9em}
\end{figure}

The sub-goal extracting module decomposes language instructions into several sub-goals, facilitating step-by-step reasoning and identifying landmarks. Benefiting from this, the agent can change its behavior adaptively during path planning, systematically explore multiple candidate nodes or sub-goals step by step, and backtrack to a specific node for re-exploration when necessary, as shown in Figure. \ref{fig:Ambiguous}.

\subsection{Wayfinding Prompt Optimization}
Wayfinding Prompt Optimization is designed to fill the gap between the perception results and the reasoning input. 

\subsubsection{High-resolution Spatial Descriptor}
While a UAV identifies landmarks, it translates their spatial data into textual form to facilitate subsequent action analysis. However, a rudimentary description of space that solely relies on the UAVs' perspective is insufficient for an LLM to accurately determine the landmarks' relative positions. For instance, a road detected in the UAVs' forward view could be centrally located or off to the side, but only knowing it's in the center prompts the UAV to follow the road. 

\begin{figure}
    \centering
    \includegraphics[width=0.9\linewidth]{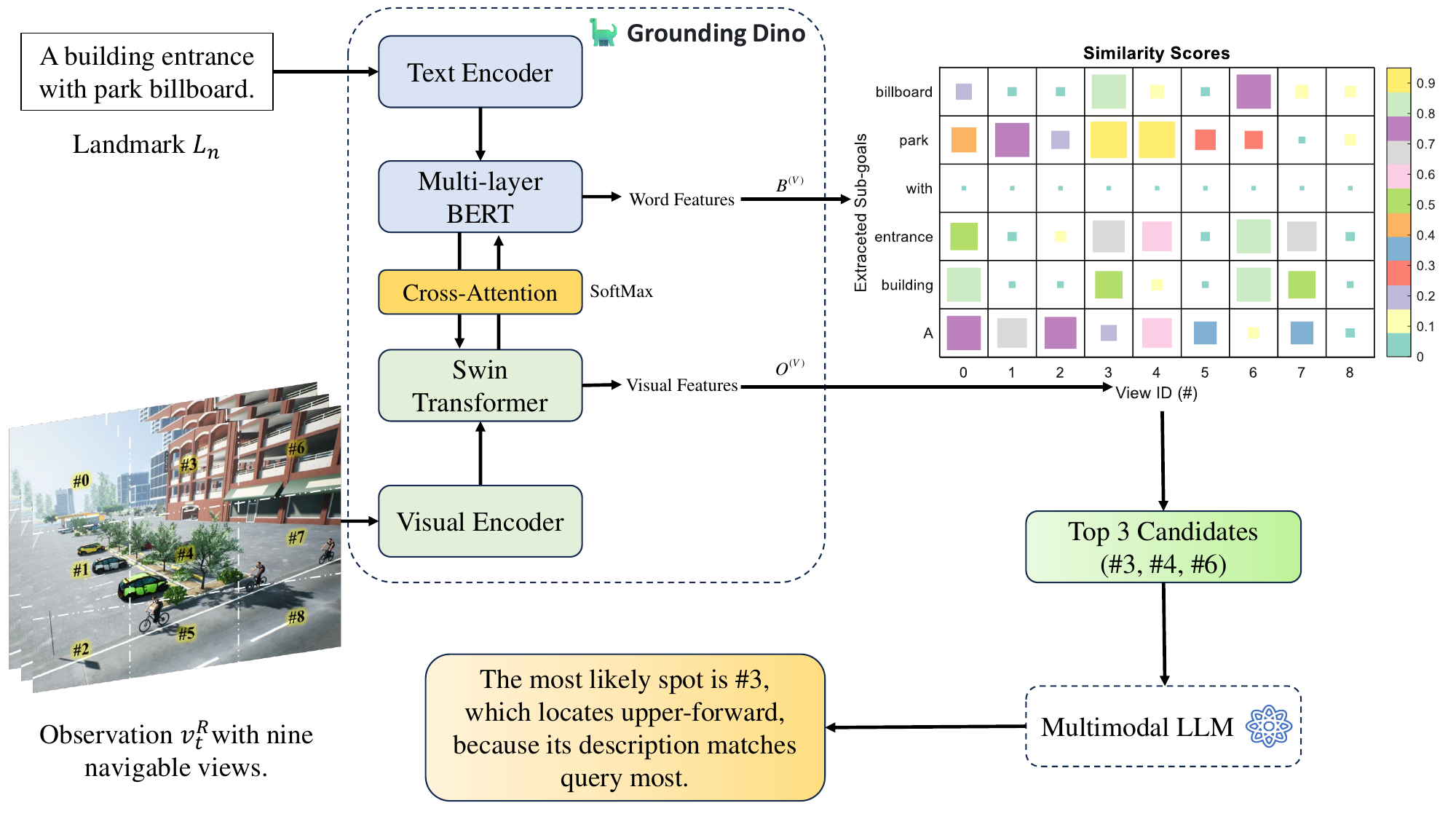}
    \caption{Schematic structure of fine-grained visual recognizer for landmark. The score matrix displays higher scores with larger blocks. The LLM evaluates and ranks query-view pairs based on the descriptive text.}
    \label{fig:Similarity}
    \vspace{-2em}
\end{figure}

To address this limitation, we introduce a high-resolution spatial descriptor (HSD). The HSD divides each perspective of the UAV into nine specific sectors, each offering a granular spatial label, such as '\#0' for the sector in the upper-left quadrant, as shown in Figure. \ref{fig:Similarity}. 
The Figure. \ref{fig:Similarity} illustrates the feature fusion process between the visual feature \(O\) extracted from the observation image and the word feature \(B\) extracted from the landmark text. For a specific inquiry denoted as ${{L}_{n}}$, the system identifies the top-K potential matches. Subsequently, the LLM crafts a descriptive text for each pairing of the inquiry and a potential match. Following this, a comprehensive assessment of these descriptive texts is conducted to ascertain the degree of similarity between the original query and each of the potential matches, thereby establishing a ranking order.
Consequently, the landmarks' spatial data is articulated not only by the UAVs' viewpoint but also by their exact coordinates within that sector.

\subsubsection{TrackBack Memory Array}
In some cases, navigation instructions may include continuous ambiguous descriptions without any landmark, such as ``turn left, then move right, then go straight", which has no objects for an LLM to refer to. The UAV fails to derive a navigable path from such ambiguous instructions, this case will directly lead to the repeated execution of a step \cite{GaoWangJingWangLi2024}. As shown in Figure. \ref{fig:Ambiguous} (left), given only the unclear commands, when an agent realizes that it has engaged in an erroneous exploration, it can only continue to explore surrounding environment aimlessly.

Consequently, we have crafted a module termed the TrackBack Memory Array (TBMA). The TBMA stores the agent’s historical trajectories and instructions in a graph, with nodes representing previously encountered landmarks and edges storing specific navigation instructions between landmarks. For any two nodes, a navigable path can be retrieved by the shortest path algorithm. Thus, the agent can derive a feasible path from the TBMA by current observed landmarks and the target landmark for ambiguous instructions. A formatted prompt is additionally designed for LLM to ask commander a clarification question depicted in Figure. \ref{fig:Illustration}. The goal of LLM is to predict the next action that is converted to a control command for the UAV to execute. The action with the highest probability is selected as the next action for the agent.
As shown in Figure. \ref{fig:Ambiguous},  incorporating history nodes enables the agent to understand spatial structures and engage in global exploration and path planning.

The TMA and HSD constitute complementary modules that operate synergistically to resolve environmental ambiguity and historically dependent instructions. When HSD detects ambiguous landmarks (e.g., "white building"), it triggers TMA to query historical references. Upon receiving temporal context from TMA, HSD refocuses on pertinent regions to re-extract features and optimizes path planning.

\subsection{Nonlinear MPC for Dynamic Obstacle Avoidance}
In Figure. \ref{fig:VLN agent} (bottom), we denote the UAV control actions by $u={{\left[ T,{{\phi }_{ref}},{{\theta }_{ref}} \right]}^{T}}$, and the corresponding UAV state vector as $x={{\left[ p,v,\phi ,\theta  \right]}^{T}}$.
In our proposed control architecture. The Airsim provides state data for the UAV \(x\) and the obstacle \(x_{\text{obstacles}}\). The obstacle data is used to classify the trajectory and generate a predicted trajectory based on the measured initial condition. The UAV state data and the obstacle trajectory is fed to the NMPC module as the solver parameter which also include \({{u}_{ref}},{{r}_{obs}},{{x}_{ref}},{{r}_{safety}}\). The NMPC generates control inputs \({{\phi }_{ref}},{{\theta }_{ref}}\) and \(T\) , which after the relevant mapping are fed to the low-level attitude controller.

The rotor aerodynamic model established in Airsim is relatively rough and its accuracy falls far short of the requirements for scientific research \cite{gao2024embodied}. If agile control of high-speed flight of drones is required, there may be significant issues when transplanting from simulation to real aircraft \cite{GaoWangJingWangLi2024}. Thus, we define the 6-DoF for the UAV by the set of Eq. \eqref{eq:kinetics}, while the full derivation of the adopted model can be found in \cite{kamel2017model}.
\begin{equation}
\left\{\begin{array}{l}
\dot{p}(t)=v(t), \\
\dot{v}(t)=R(\phi, \theta)\left[\begin{array}{c}
0 \\
0 \\
T
\end{array}\right]+\left[\begin{array}{c}
0 \\
0 \\
-g
\end{array}\right]-A v(t), \\
A=\left[\begin{array}{ccc}
A_x & 0 & 0 \\
0 & A_y & 0 \\
0 & 0 & A_z
\end{array}\right], \\
\dot{\phi}(t)=\frac{1}{\tau_\phi}\left(K_\phi \phi_{\mathrm{ref}}(t)-\phi(t)\right), \\
\dot{\theta}(t)=\frac{1}{\tau_\theta}\left(K_\theta \theta_{\mathrm{ref}}(t)-\theta(t)\right) ,
\end{array}\right. \label{eq:kinetics}
\end{equation}
where the position of the UAV is denoted by $p(t)={{\left[ x,y,z \right]}^{T}}$ and its linear velocity in the global frame of reference by $v(t)={{\left[ u,v,w \right]}^{T}}$. The roll and pitch angles, \(\theta\) and \(\phi\), both within the range \(-\pi,\pi\), are defined along the \(x^W\) and \(y^W\) axes,respectively. The rotation matrix $R\left( \theta \left( t \right),\phi \left( t \right) \right)\in SO\left( 3 \right)$ describes the UAVs attitude in Euler form. The reference inputs to the system are \({{\theta }_{ref}},{{\phi }_{ref}}\in R\), and \(T\geq 0\), representing the desired roll, pitch, and total thrust, respectively. The model assumes that the acceleration is influenced solely by the magnitude and angle of the thrust vector generated by the motors, along with linear damping terms ${{A}_{x}},{{A}_{y}},{{A}_{z}}\in R$ and the gravitational acceleration \(g\). The attitude dynamics are modeled as first-order systems with gains \({{K}_{\theta }}, {{K}_{\phi }}\in R\) and time constants \({{\tau }_{\theta }}, {{\tau }_{\phi }}\in R\), capturing the closed-loop behavior of a low-level controller. This implies that the UAV is equipped with a lower-level attitude controller that translates thrust, roll, and pitch commands into motor commands.

The overall NMPC algorithm flow is shown in Algorithm \ref{Algo:MPC}, which could be divided into four parts:

\normalem
\begin{algorithm}[t]
\caption{Nonlinear MPC for Collision Avoidance and Control of UAVs with Dynamic Obstacles}
\label{Algo:MPC}
\DontPrintSemicolon
\KwIn{Initial state of UAV, $ x_{\text{UAV}}(0) $, initial states of dynamic obstacles, $ x_{\text{obstacles}}(0) $, prediction horizon, $ T_p $, control horizon, $ T_c $, and safety constraints, $ C $.}
\KwOut{Optimal control inputs, $ u^* $, for UAV to avoid collisions with dynamic obstacles.}

\SetKw{SetCurrentTime}{}
\SetKw{SetCurrentStateUAV}{}
\SetKw{SetCurrentStateObstacles}{}

\SetCurrentTime $t = 0$\;
\SetCurrentStateUAV $x_{\text{UAV}}(t) = x_{\text{UAV}}(0)$\;
\SetCurrentStateObstacles $x_{\text{obstacles}}(t) = x_{\text{obstacles}}(0)$\;

\While{UAV is operational}{
    Measure $x_{\text{UAV}}(t)$ and $x_{\text{obstacles}}(t)$\;
    Predict $ \hat{x}_{\text{UAV}}(t+1) $ and $ \hat{x}_{\text{obstacles}}(t+1) $ over $ T_p $\;
    \For{each obstacle $i$ in $x_{\text{obstacles}}(t)$}{
        Predict $ \hat{x}_i(t+1) $ over $ T_p $\;
    }
    Formulate the optimization problem:
    \begin{align*}
        \min_{u} \quad & J(x_{\text{UAV}}, u, T_c) \\
        \text{s.t.} \quad & x_{\text{UAV}}(k+1) = f(x_{\text{UAV}}(k), u(k)), \\
        & \quad k = 0, \ldots, T_c-1 \\
        & \quad \text{collision avoidance constraints}, \\
        & \quad \text{actuator limits}.
    \end{align*}
    Solve for $ u^* $\;
    Apply $ u^*(0) $ to UAV\;
    Update $ t = t + \Delta t $\;
    \If{UAV has reached destination or emergency stop}{
        \textbf{break}
    }
}

\end{algorithm}
\vspace{1em}

\subsubsection{Cost Function}

By assigning a cost to the alignment of current and predicted states and inputs, a nonlinear optimizer is charged with determining the optimal sequence of control actions, which corresponds to the minimum value of the \textit{cost function}:
\begin{equation}
\begin{aligned}
& J\left(\boldsymbol{x}_k, \boldsymbol{u}_k, u_{k-1 \mid k}\right)=\sum_{j=0}^N \underbrace{\left\|x_{\text {ref }}-x_{k+j \mid k}\right\|_{Q_x}^2}_{\text {State cost }} \\
& \quad+\underbrace{\left\|u_{\mathrm{ref}}-u_{k+j \mid k}\right\|_{Q_u}^2}_{\text {Input cost }}+\underbrace{\left\|u_{k+j \mid k}-u_{k+j-1 \mid k}\right\|_{Q_{\Delta u}}^2}_{\text {Input smoothness cost }},
\end{aligned}
\end{equation}
Where 
\( J \) is the cost function.
\( x_k \) is the state vector at time step \( k \).
\( u_k \) is the control action vector at time step \( k \).
\( u_{k-1|k} \) is the control action vector at the previous time step.
\( x_{\text{ref}} \) is the reference state vector.
\( u_{\text{ref}} \) is the reference control input vector.
\( x_{k+j|k} \) is the predicted state at time step \( k+j \), produced at the time step \( k \).
\( u_{k+j|k} \) is the control action applied at time step \( k+j \), predicted at time step \( k \).
\( Q_x \), \( Q_u \), and \( Q_{\Delta u} \) are positive definite weight matrices for the states, inputs, and input rates, respectively.
\( N \) is the prediction horizon.

\subsubsection{Obstacle Definition}
We utilize a spherical obstacle model to represent dynamic obstacles that may interact with the UAV. The obstacle is defined with an arbitrary trajectory, where the position of the sphere at each time step in the prediction horizon is set as an input to the solver. The obstacle constraint is formulated using the function  $[h]^+ = \max\{0, h\}$ , which allows expressing a constrained area by choosing  $h$  as an expression that is positive inside the constrained area and negative outside of it.

For a spherical obstacle with radius  $r_{\text{obs}}$  and an additional safety radius  $r_s$ , the obstacle positions \(p_{\text{obs}}\) are the world-frame coordinates of the center of the sphere. The constraint ensuring the UAV in position \(p\) does not enter the obstacle's space is given by:
\begin{equation}
\begin{array}{r}
h_{\text {sphere }}\left(p, \xi_{\mathrm{obs}}\right)=\left[\left(r_{\mathrm{obs}}+r_s\right)^2-\left(x-x_{\mathrm{obs}}\right)^2-\right. \\
\left(y-y_{\mathrm{obs}}\right)^2-\left(z-z_{\mathrm{obs}}\right)^2]_+=0,
\end{array}
\end{equation}
where  $\xi_{\text{obs}} = [r_{\text{obs}}, r_s, p_{\text{obs}}]$, $p_{\text{obs}} = [x_{\text{obs}}, y_{\text{obs}}, z_{\text{obs}}]$. This constraint is extended to a dynamic obstacle by forming a separate constraint for each predicted time step, thus fully describing and parametrizing the obstacle's trajectory by  $\xi_{\text{obs}}$.

\subsubsection{Input Constraints}
To ensure smooth and stable control actions, particularly when avoiding dynamic obstacles, we impose constraints on the rate of change of the control inputs. 
\begin{equation}
\left\{\begin{array}{l}
\left|\phi_{r e f, k+j-1 \mid k}-\phi_{r e f, k+j \mid k}\right| \leq \Delta \phi_{\max }, \\
\left|\theta_{r e f, k+j-1 \mid k}-\theta_{r e f, k+j \mid k}\right| \leq \Delta \theta_{\max },
\end{array}\right.
\end{equation}
where \(\Delta \phi_{\max }\) and \(\Delta \theta_{\max }\) denote the maximum change in input per time step. 

\subsubsection{Real-time Optimization}

The NMPC problem is solved using the Proximal Averaged Newton for Optimal Control (PANOC) algorithm and its associated software OpEn (Optimization Engine) \cite{OPEN}. PANOC is a Newton-type method specifically designed for optimal control problems, known for its low memory and computational load, satisfying low latency for UAVs.

The optimization problem is formulated as:
\begin{equation}
\begin{array}{r}
 \text{Minimize} \quad z \in Z \, f(z, \rho), \\

 \text{subject to:} \quad F(z, \rho) = 0,
\end{array}
\end{equation}
where  $f$  is a Lipschitz-differentiable function,  $F$  is a vector-valued mapping,  $z$  is the decision variable, and  $\rho$  includes initial conditions, references, and the obstacle trajectory. The equality constraints are handled using a quadratic penalty method, which gradually moves the cost-minima until none of the constraints are violated or until a specified tolerance is met.

\section{The 3D Experimental Platform} \label{sec:Platform}
The core of this platform is a high-fidelity 3D environment in order to further validate the UAVs' VLN tasks. In this environment, we have established a pipeline for agents to be deployed in the environment, read input with first-view observations, and make decisions.

\subsection{3D Environment} 
The basic environment includes 3D models of buildings, streets, and other outdoor entities. Buildings include shopping malls, residential complexes, and public facilities. 
The streets are modeled to include all necessary components such as lanes, intersections, traffic signals, and road markings.
Vehicles and Pedestrians \cite{zhangMirage2022} provides realistic interactions and behaviors that mimic real-world traffic and pedestrian dynamics.
Other elements include Street furniture (benches, streetlights, signs), vegetation (trees, shrubs, lawns), and urban amenities (bus stops, entrances, public restrooms).

\begin{figure}
    \centering
    \includegraphics[width=\linewidth]{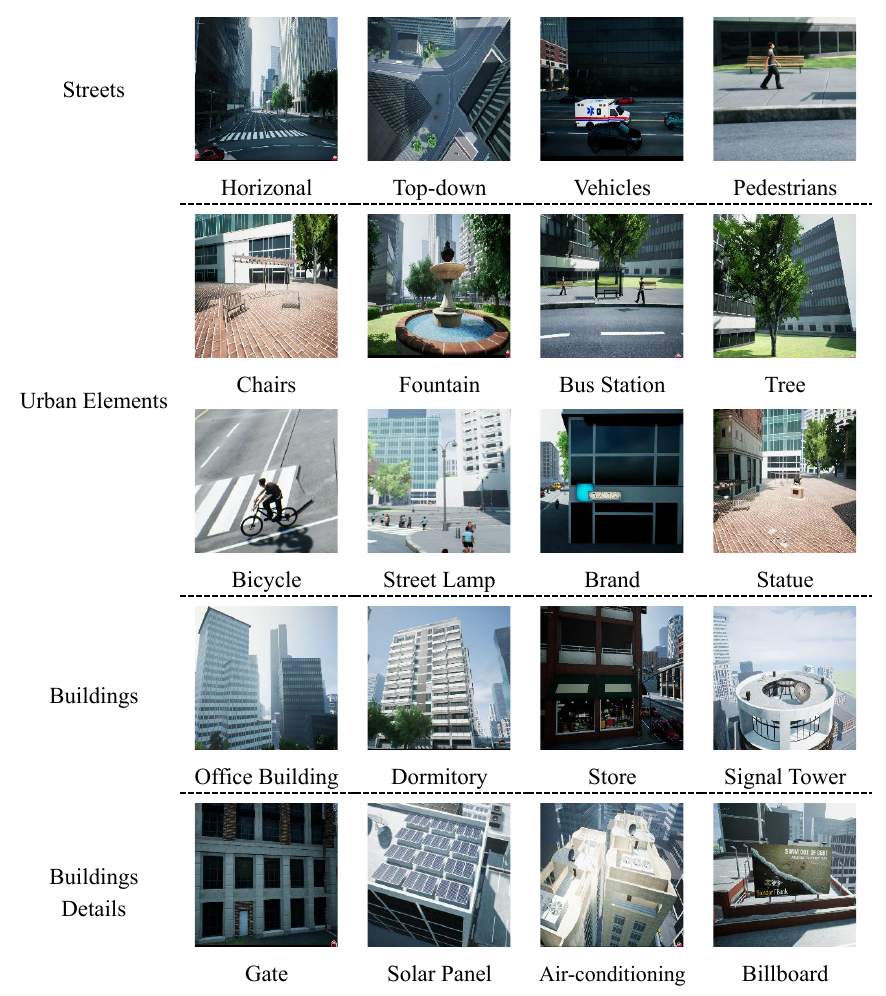}
    \caption{The image showcases various components of our city simulator. Diverse details are added on the roof and exterior walls of high-rise buildings for aerial agents recognize conveniently.}
    \label{fig:image showcases}
    \vspace{-1em}
\end{figure}

We offer an in-depth look at the environment depicted in Figure. \ref{fig:image showcases}. It stands out among its peers by delivering high-fidelity 3D models and seamless real-time data integration. Notably, it includes specialized support for drone simulations, positioning it as an essential tool for the development and refinement of autonomous systems in urban contexts, focusing on perception, navigation, and strategic planning.

\subsection{VLN Simulator} 
The simulator is developed based on AirSim \cite{shah2017airsimhighfidelityvisualphysical} and Unreal Engine 4 \cite{UE4}. Below we detail its visual perceptions and action space. 

\subsubsection{Visual Observations}
In the simulator, an embodied agent can move and observe in the continuous outdoor environment freely. At each step \(t\), the simulator can output an RGB image \(v_{t}^{R}\), a depth image $v_{t}^{D}$ and a semantic segments image $v_{t}^{S}$ of its front view.

\subsubsection{Action Space}
Motion control involves setting target positions \({{\left[ x,y,z \right]}^{T}}\), target velocities \({{\left[ u,v,w \right]}^{T}}\), and target orientations \({[\theta, \phi, \psi]^{T}}\). Camera control allows for viewpoint adjustments, and other controls include starting or stopping the drone’s flight.
Unlike TouchDown\cite{chen2019touchdown}, R2R\cite{anderson2018vision}, RxR\cite{ku2020room}, our simulation environment is a continuous space so that the simulated drone can fly in the sky continuously to any point within the environment.

\section{Experiment and Results} \label{sec:Experiment}

\subsection{Experiment Setup}
\begin{figure}  
\centering
\subfigure[Distribution of trajectory length.]{
\begin{minipage}[b]{0.24\textwidth}
\includegraphics[width=1\textwidth]{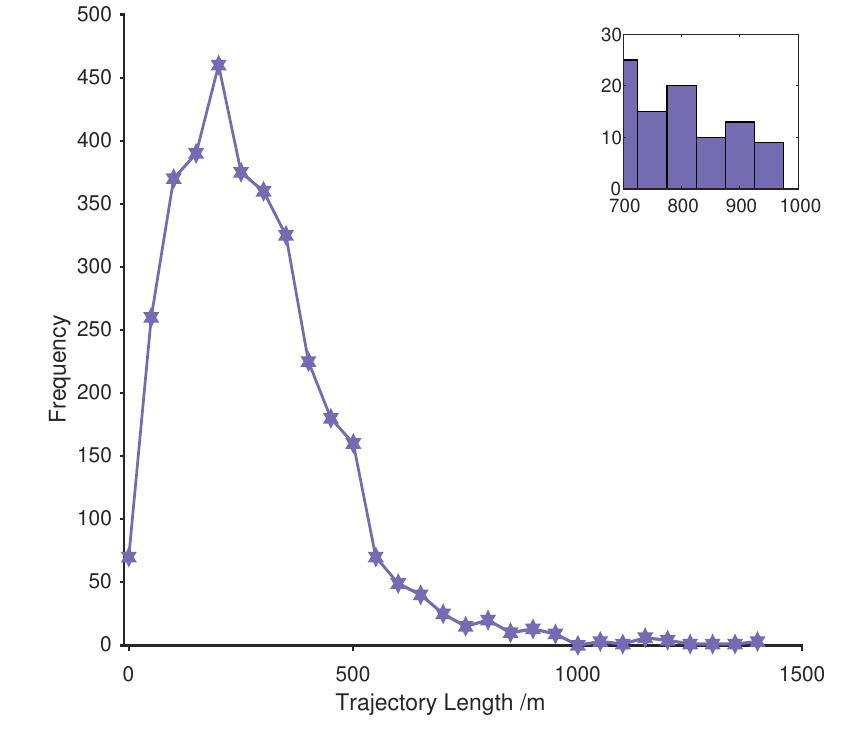} 
\end{minipage}
}
\hfill
\subfigure[Word cloud of common terms in AVDN dataset.]{
\begin{minipage}[b]{0.2\textwidth}
\includegraphics[width=1\textwidth]{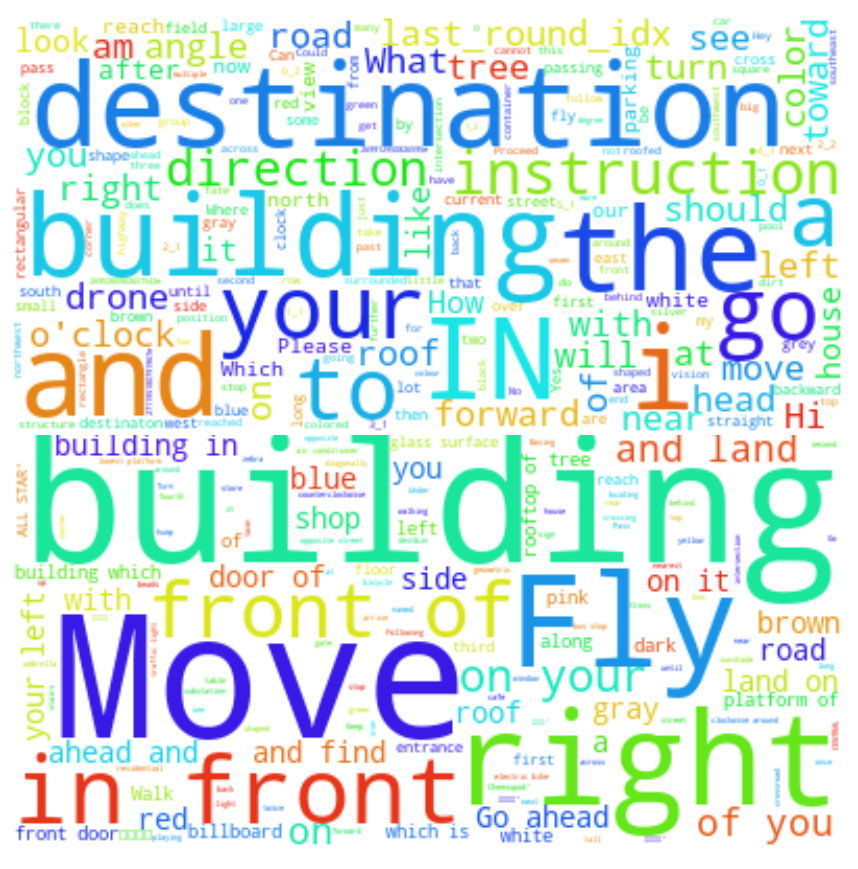} 
\end{minipage}
}
\caption{UAV trajectory length of the AVDN dataset and word clouds of frequent words that appear in the dialogs.}
\label{fig:Statistics}
\vspace{-1em}
\end{figure}

We conduct experiments using a novel and complex dataset provided by \cite{FanChenJiangZhouZhangWang2023}, which we refer to as the ``AVDN" dataset, to assess the performance of different component functions of our system.
The distribution of trajectory length and the word cloud are presented in Figure. \ref{fig:Statistics}. As shown in Figure. \ref{fig:Statistics} (a), the average path length is \(287m\). The most frequent instruction word are shown in Figure. \ref{fig:Statistics} (b).

Finally, We evaluated the algorithm on Airsim and UE4 with GPT-4, on a laptop with an Intel i9 12th generation CPU and an NVIDIA GeForce RTX 4070 GPU. Considering the setting of the outdoor environment, the depth sensor is allowed to perceive 100 meters ahead. We set the field of view of camera sensor as 90 degrees.

\subsection{Quantitative Results}

Table \ref{tab:Results_1} presents a comparative analysis of various models on the AVDN dataset, focusing on their performance in both seen validation and unseen test conditions. The metrics of interest are SPL (Success weighted by Path Length) and SR (Success Rate), both of which are critical for evaluating the navigational capabilities of UAVs in vision-language navigation tasks.
The baseline methods include 'Random', Cross Modal Alignment ('CMA'), 'Seq2Seq' and the 'NavGPT'.
The 'NavGPT' model shows a significant improvement over the other baseline. However, it underperforms in the unseen test set, with an SPL of 18.9\% and an SR of 16.6\%. The models 'w MPC', and 'w NMPC' show varying degrees of improvement. The 'w MPC' model, in particular, shows a notable performance in the seen validation phase, with an SPL of 14.34\% and an SR of 17.3\%. However, it underperforms in the unseen test set, with an SPL of 12.9\% and an SR of 15.17\%. The 'w/o HSD' model exhibits a notable performance, especially in the unseen validation phase. The 'w/o TMA' model shows a competitive performance, particularly in the unseen test set. Notably, our 'Ours Full' model, which incorporates all the components, outperforms all other models, achieving the highest scores in both seen and unseen conditions. In the unseen test set, it notably surpasses the others with an SPL of 28.11\% and an SR of 42.37\%, indicating its robustness and generalizability to new environments. These results underscore the effectiveness of our full model in handling the complexities of vision-language navigation for UAVs.

\begin{figure}
    \centering
    \includegraphics[width=1\linewidth]{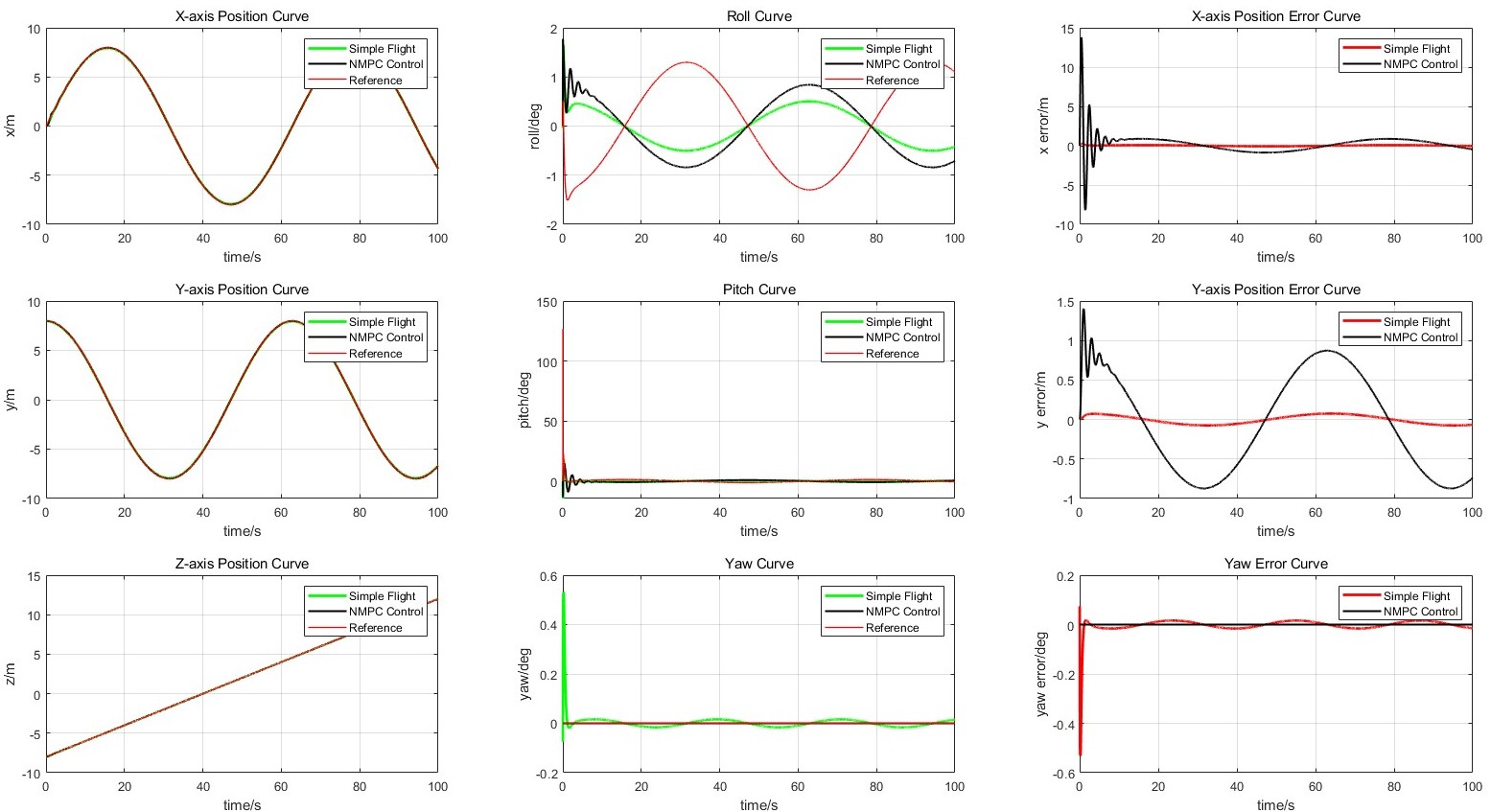}
    \caption{6-DOF curve and error curve of UAV trajectory in static obstacle case.}
    \label{fig:static}
    \vspace{-1em}
\end{figure}

\begin{figure}[ht!]
    \centering
    \includegraphics[width=1\linewidth]{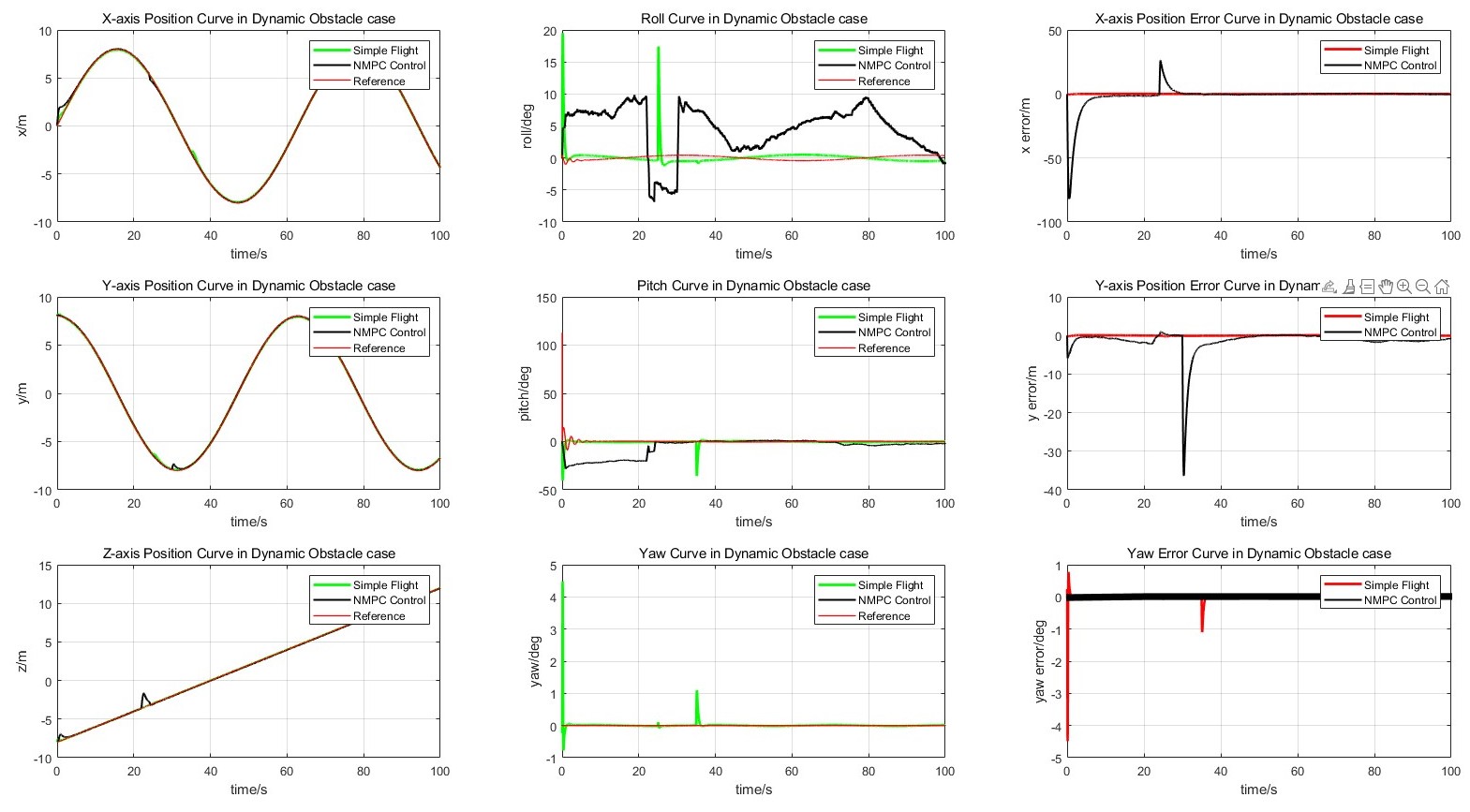}
    \caption{6-DOF curve and error curve of UAV trajectory in additional dynamic obstacle case.}
    \label{fig:dynamic}
    \vspace{-2em}
\end{figure}

\begin{table}[]
\centering
\caption{Comparison on the validation and unseen test set of the AVDN dataset.}
\label{tab:Results_1}
\resizebox{\columnwidth}{!}{%
\begin{tabular}{ccccccc}

\toprule
\multirow{3}{*}{\textbf{Model}} & \multicolumn{6}{c}{\textbf{AVDN}}                                                                      \\
 & \multicolumn{2}{c}{\textbf{Seen Validation}} & \multicolumn{2}{c}{\textbf{Unseen Validation}} & \multicolumn{2}{c}{\textbf{Unseen Testing}} \\
                                & \textbf{SPL/\% \(\uparrow\)} & \textbf{SR/\% \(\uparrow\)} & \textbf{SPL/\%} & \textbf{SR/\% \(\uparrow\)} & \textbf{SPL/\% \(\uparrow\)} & \textbf{SR/\% \(\uparrow\)} \\  \hline
Random                          & 0.5             & 1.6            & 0.2             & 1              & 0.5             & 1.1            \\
CMA & 8.2             & 10.5            & 12.8             & 6.7              & 8.3             & 9.7           \\
Seq2Seq\cite{8578485} & 5.1             & 6.4            & 7.9	             & 4.2              & 5.6             & 5.1           \\
NavGPT\cite{Zhou2023NavGPTER} & 12.3             & 14.6            & 15.2		             & 10.8              & 18.9             & 16.6 \\ \hline
w PID                           &6.8              &8.7             &9.4             &5.3	               &6.9              &12.9             \\
w MPC                        & {\uline {14.34}}      & \textbf{17.3}  & 16.5            & {\uline{20.4}}     & 12.9            & {15.17}     \\
w NMPC  &13.9   &{\uline{16.2}}  &{\uline{17.1}}  &18.5  &{\uline{22.4}}	 &{\uline{26.8}}
	\\
w/o HSD                         & 11.6            & 13       & \textbf{18.3}   & 20             & 12.6            & 14.1           \\
w/o TMA                         & 11.57           & 12.97          & \textbf{18.3}   & 19.95          & 18.17     & 15.32      \\
\textbf{Ours Full}                           & \textbf{14.7}  & \textbf{17.3}  & 16.62     & \textbf{20.44} & \textbf{28.11}  & \textbf{42.37} \\ 
\bottomrule
\end{tabular}%
}
\end{table}
The Figure. \ref{fig:static} and Figure. \ref{fig:dynamic} illustrate the position curves for the UAV along the X, Y, and Z axes under two control methods: Simple Flight and NMPC. The Simple Flight controller in AirSim uses a PID (Proportional Integral Derivative) control algorithm. The NMPC Control, represented by the black line, closely follows the reference trajectory, depicted in red, with minimal deviation. This is particularly evident in the Z-axis Position Curve, where the NMPC Control maintains a steady ascent, closely mirroring the reference, unlike the Simple Flight control, which exhibits a more erratic pattern. 

With the black line adhering closely to the reference in the Roll, Pitch, and Yaw curves, indicating precise control over the UAVs' orientation. Position error curves quantify the deviation from the desired trajectory. For the NMPC Control, the error curves are significantly lower than those for Simple Flight, especially in the X and Y axes, suggesting that NMPC is more effective in maintaining the UAVs' position within the designated path. The Yaw Error Curve also shows a more stable performance by NMPC, with errors quickly dampening to near zero, whereas Simple Flight exhibits a more pronounced and sustained error. These analyses suggest that NMPC is a more reliable control strategy for UAVs in urban environments, offering better obstacle avoidance and trajectory tracking.

\subsection{Qualitative Analysis} 
\begin{figure}
    \centering
    \includegraphics[width=0.95\linewidth]{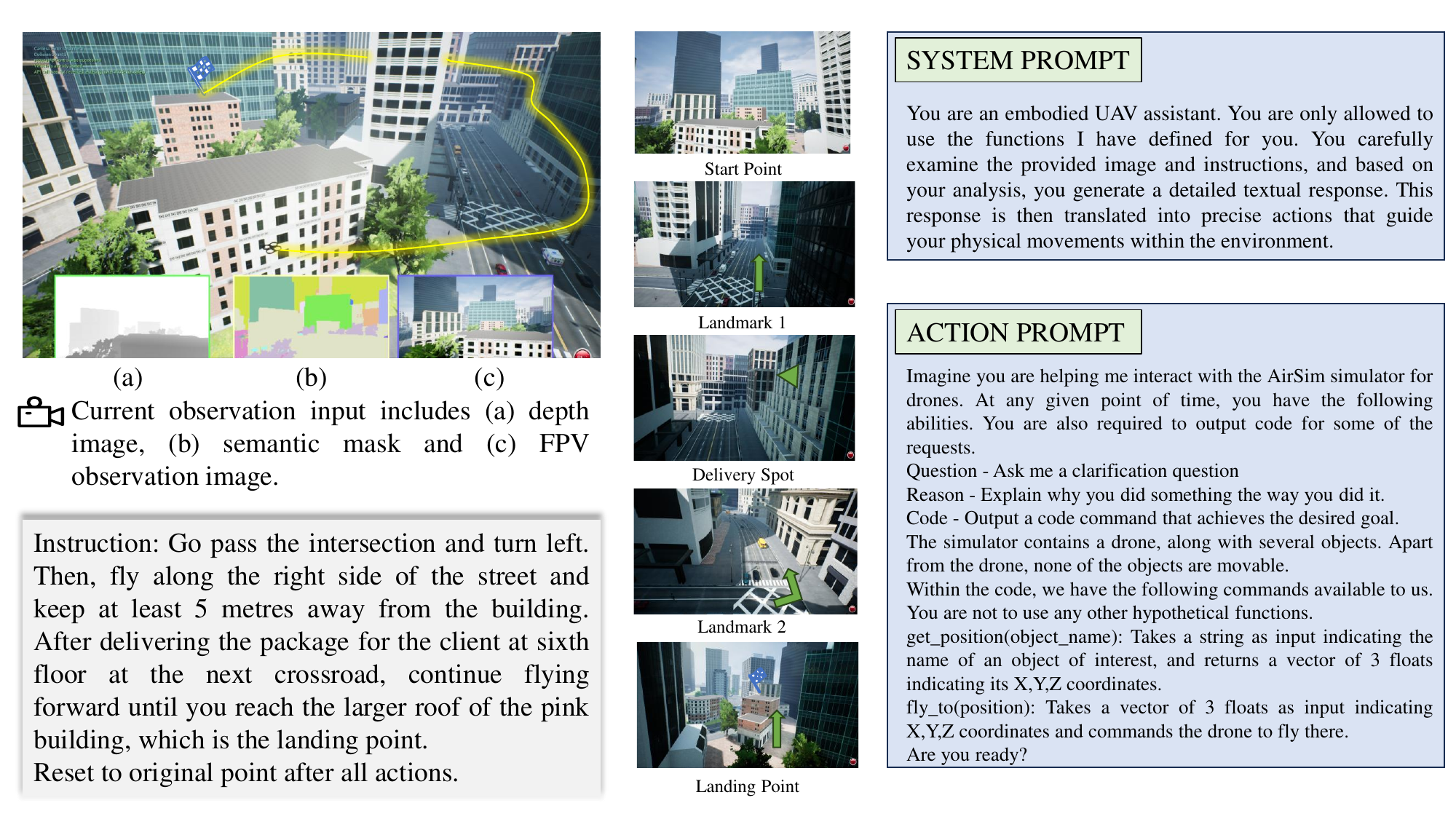}
    \caption{An illustration of the embodied VLN tasks in city environment. The yellow line shows the agent’s ground truth trajectory, and the chequered flag represents the end of it.}
    \label{fig:Illustration}
    \vspace{-1em}
\end{figure}

Figure. \ref{fig:Illustration} shows a qualitative analysis. An intelligent agent is required to distinguish referred buildings/objects by their spatial relationship from a bird-eye view. A system prompt is leveraged to describe the embodiment, an action prompt to provide feedback, the potential actions, and the output instruction.

As shown in Figure. \ref{fig:Illustration}, the proposed SkyVLN task requires the intelligent agent  to fly to the destination by following a given natural language instruction and its first-person view visual perceptions provided by the camera. 

\subsection{Ablation Study} 

Table \ref{tab:Results_2} provides an ablation study comparing the performance of different LLMs in the context of Unseen Testing. The GPT-4V model shows a balanced performance with an SPL of 16.62\% and an SR of 34.9\%, while maintaining a relatively low NE of 62.35, indicating efficient navigation with a good success rate. In contrast, the GPT-4o model achieves the highest SPL of 34.25\%, suggesting a strong ability to follow the path to success, but it has a lower SR of 20.44\% and a higher NE of 90.11, which may imply a less reliable success rate and less efficient navigation. The GPT-4 Turbo model presents a moderate SPL of 25.12\% and the lowest SR of 15.62\%, coupled with the highest NE of 127.87, suggesting that while it explores more, its success rate and path efficiency are not as optimal as the other models. 

\begin{table}[]
\centering
\caption{Ablation study for the different LLMs.}
\label{tab:Results_2}
\resizebox{0.7\columnwidth}{!}{%
\begin{tabular}{cccc}
\toprule
\multirow{2}{*}{\textbf{LLM}} & \multicolumn{3}{c}{\textbf{Unseen Testing}}       \\
                              & \textbf{SPL/\%} & \textbf{SR/\%} & \textbf{NE/m}  \\ \hline
GPT-4V                        & 16.62           & \textbf{34.9}  & \textbf{62.35} \\
GPT-4o                        & \textbf{34.25}  & \uline{20.44}          & \uline{90.11}          \\
GPT-4 Turbo                   & \uline{25.12}           & 15.62          & 127.87         \\ 
\bottomrule
\end{tabular}%
}
\end{table}

\section{Conclusion} \label{sec:conclusion}
The proposed SkyVLN represents an innovative vision-and-language navigation framework, meticulously designed for UAV operations within intricate urban environments. This approach is anticipated to substantially alleviate the cognitive demands placed on operators, thereby facilitating more efficient task execution across a variety of contexts. Moreover, the integration of a historical path memory feature allows the UAV to preserve contextual awareness over time, which is paramount for navigating the dynamic intricacies of complex urban landscapes. The research undertaken exploits multimodal perception strategies to enhance the efficacy and transparency of reasoning and decision-making processes through the employment of LLMs. It also incorporates nonlinear model predictive control tailored for dynamic scenarios and assembles a highly detailed 3D simulation environment to rigorously substantiate the resilience of navigation control systems.

Future research should focus on optimizing the balance between safety and mission efficiency, expanding to coordinated multi-UAV operations by integrating swarm intelligence, and further enhancing the applicability of UAV control to severe environmental conditions. The findings of this study pave the way for more sophisticated and intelligent UAV navigation, highlighting the potential of LLMs to revolutionize autonomous UAV operations.

\section{Acknowledgement} \label{sec:Acknowledgement}
This work was partially supported by the National Natural Science Foundation of China Grant 62373315, 62403401 and U24A20252, and the Nansha Key Science and Technology Project 2023ZD006.

\normalem
\bibliographystyle{IEEEtran}
\bibliography{Arxiv_IROS2025}

\begin{thebibliography}{10}
\providecommand{\url}[1]{#1}
\csname url@rmstyle\endcsname
\providecommand{\newblock}{\relax}
\providecommand{\bibinfo}[2]{#2}
\providecommand\BIBentrySTDinterwordspacing{\spaceskip=0pt\relax}
\providecommand\BIBentryALTinterwordstretchfactor{4}
\providecommand\BIBentryALTinterwordspacing{\spaceskip=\fontdimen2\font plus
\BIBentryALTinterwordstretchfactor\fontdimen3\font minus \fontdimen4\font\relax}
\providecommand\BIBforeignlanguage[2]{{%
\expandafter\ifx\csname l@#1\endcsname\relax
\typeout{** WARNING: IEEEtran.bst: No hyphenation pattern has been}%
\typeout{** loaded for the language `#1'. Using the pattern for}%
\typeout{** the default language instead.}%
\else
\language=\csname l@#1\endcsname
\fi
#2}}

\bibitem{li2021networked}
X.~Li and A.~V. Savkin, ``Networked unmanned aerial vehicles for surveillance and monitoring: A survey,'' \emph{Future Internet}, vol.~13, no.~7, p. 174, 2021.

\bibitem{tian2025uavsmeetllmsoverviews}
\BIBentryALTinterwordspacing
Y.~Tian, F.~Lin, Y.~Li, T.~Zhang, Q.~Zhang, X.~Fu, J.~Huang, X.~Dai, Y.~Wang, C.~Tian, B.~Li, Y.~Lv, L.~Kovács, and F.-Y. Wang, ``Uavs meet llms: Overviews and perspectives toward agentic low-altitude mobility,'' 2025. [Online]. Available: \url{https://arxiv.org/abs/2501.02341}
\BIBentrySTDinterwordspacing

\bibitem{al2020comprehensive}
F.~Al-Turjman and H.~Zahmatkesh, ``A comprehensive review on the use of ai in uav communications: Enabling technologies, applications, and challenges,'' \emph{Unmanned Aerial Vehicles in Smart Cities}, pp. 1--26, 2020.

\bibitem{rasley2020deepspeed}
J.~Rasley, S.~Rajbhandari, O.~Ruwase, and Y.~He, ``Deepspeed: System optimizations enable training deep learning models with over 100 billion parameters,'' in \emph{Proceedings of the 26th ACM SIGKDD International Conference on Knowledge Discovery and Data Mining}, 2020, pp. 3505--3506.

\bibitem{kurunathan2023machine}
H.~Kurunathan, H.~Huang, K.~Li, W.~Ni, and E.~Hossain, ``Machine learning-aided operations and communications of unmanned aerial vehicles: A contemporary survey,'' \emph{IEEE Communications Surveys and Tutorials}, 2023.

\bibitem{Zhou2023NavGPTER}
\BIBentryALTinterwordspacing
G.~Zhou, Y.~Hong, and Q.~Wu, ``Navgpt: Explicit reasoning in vision-and-language navigation with large language models,'' in \emph{AAAI Conference on Artificial Intelligence}, 2023. [Online]. Available: \url{https://api.semanticscholar.org/CorpusID:258947250}
\BIBentrySTDinterwordspacing

\bibitem{Zhou2024NavGPT2UN}
\BIBentryALTinterwordspacing
G.~Zhou, Y.~Hong, Z.~Wang, X.~E. Wang, and Q.~Wu, ``Navgpt-2: Unleashing navigational reasoning capability for large vision-language models,'' in \emph{European Conference on Computer Vision}, 2024. [Online]. Available: \url{https://api.semanticscholar.org/CorpusID:271244657}
\BIBentrySTDinterwordspacing

\bibitem{park2023visual}
S.-M. Park and Y.-G. Kim, ``Visual language navigation: A survey and open challenges,'' \emph{Artificial Intelligence Review}, vol.~56, no.~1, pp. 365--427, 2023.

\bibitem{LyuMultimodalLLMsMeetPlaceRecognition2024}
\BIBentryALTinterwordspacing
Z.~Lyu, J.~Zhang, M.~Lu, Y.~Li, and C.~Feng, ``\BIBforeignlanguage{en}{Tell me where you are: Multimodal llms meet place recognition},'' \emph{\BIBforeignlanguage{en}{arXiv preprint arXiv:2406.17520}}, no. arXiv:2406.17520, June 2024, arXiv:2406.17520 [cs]. [Online]. Available: \url{http://arxiv.org/abs/2406.17520}
\BIBentrySTDinterwordspacing

\bibitem{chen2019touchdown}
H.~Chen, A.~Suhr, D.~Misra, N.~Snavely, and Y.~Artzi, ``Touchdown: Natural language navigation and spatial reasoning in visual street environments,'' in \emph{Proceedings of the IEEE/CVF Conference on Computer Vision and Pattern Recognition}, 2019, pp. 12\,538--12\,547.

\bibitem{KrantzBeyondtheNavGraph2020}
J.~Krantz, E.~Wijmans, A.~Majumdar, D.~Batra, and S.~Lee, ``Beyond the nav-graph: Vision-and-language navigation in continuous environments,'' in \emph{Computer Vision -- ECCV 2020}, A.~Vedaldi, H.~Bischof, T.~Brox, and J.-M. Frahm, Eds.\hskip 1em plus 0.5em minus 0.4em\relax Cham: Springer International Publishing, 2020, pp. 104--120.

\bibitem{ku2020roomacrossroommultilingualvisionandlanguagenavigation}
\BIBentryALTinterwordspacing
A.~Ku, P.~Anderson, R.~Patel, E.~Ie, and J.~Baldridge, ``Room-across-room: Multilingual vision-and-language navigation with dense spatiotemporal grounding,'' 2020. [Online]. Available: \url{https://arxiv.org/abs/2010.07954}
\BIBentrySTDinterwordspacing

\bibitem{Hashim_2025}
\BIBentryALTinterwordspacing
H.~A. Hashim, ``Advances in uav avionics systems architecture, classification and integration: A comprehensive review and future perspectives,'' \emph{Results in Engineering}, vol.~25, p. 103786, Mar. 2025. [Online]. Available: \url{http://dx.doi.org/10.1016/j.rineng.2024.103786}
\BIBentrySTDinterwordspacing

\bibitem{10294277}
B.-B. Hu, H.-T. Zhang, B.~Liu, J.~Ding, Y.~Xu, C.~Luo, and H.~Cao, ``Coordinated navigation control of cross-domain unmanned systems via guiding vector fields,'' \emph{IEEE Transactions on Control Systems Technology}, vol.~32, no.~2, pp. 550--563, 2024.

\bibitem{wang2024visionbaseddeepreinforcementlearning}
\BIBentryALTinterwordspacing
J.~Wang, Z.~Yu, D.~Zhou, J.~Shi, and R.~Deng, ``Vision-based deep reinforcement learning of uav autonomous navigation using privileged information,'' 2024. [Online]. Available: \url{https://arxiv.org/abs/2412.06313}
\BIBentrySTDinterwordspacing

\bibitem{choutri2022multi}
K.~Choutri, M.~Lagha, S.~Meshoul, M.~Batouche, Y.~Kacel, and N.~Mebarkia, ``A multi-lingual speech recognition-based framework to human-drone interaction,'' \emph{Electronics}, vol.~11, no.~12, p. 1829, 2022.

\bibitem{tagliabue2023real}
A.~Tagliabue, K.~Kondo, T.~Zhao, M.~Peterson, C.~T. Tewari, and J.~P. How, ``Real: Resilience and adaptation using large language models on autonomous aerial robots,'' \emph{arXiv preprint arXiv:2311.01403}, 2023.

\bibitem{wang2024realisticuavvisionlanguagenavigation}
\BIBentryALTinterwordspacing
X.~Wang, D.~Yang, Z.~Wang, H.~Kwan, J.~Chen, W.~Wu, H.~Li, Y.~Liao, and S.~Liu, ``Towards realistic uav vision-language navigation: Platform, benchmark, and methodology,'' 2024. [Online]. Available: \url{https://arxiv.org/abs/2410.07087}
\BIBentrySTDinterwordspacing

\bibitem{devlin2018bert}
J.~Devlin, M.-W. Chang, K.~Lee, and K.~Toutanova, ``Bert: Pre-training of deep bidirectional transformers for language understanding,'' \emph{arXiv preprint arXiv:1810.04805}, 2018.

\bibitem{luo2024language}
S.~Luo, Y.~Yao, H.~Zhao, and L.~Song, ``A language model-based fine-grained address resolution framework in uav delivery system,'' \emph{IEEE Journal of Selected Topics in Signal Processing}, 2024.

\bibitem{10817801}
J.~Lai, Z.~Wu, Z.~Ren, Q.~Tan, and H.~Xiao, ``Optimal navigation of an automatic guided vehicle with obstacle constraints: A broad learning-based approach,'' \emph{IEEE Transactions on Emerging Topics in Computational Intelligence}, pp. 1--15, 2024.

\bibitem{GYAGENDA2022104069}
\BIBentryALTinterwordspacing
N.~Gyagenda, J.~V. Hatilima, H.~Roth, and V.~Zhmud, ``A review of gnss-independent uav navigation techniques,'' \emph{Robotics and Autonomous Systems}, vol. 152, p. 104069, 2022. [Online]. Available: \url{https://www.sciencedirect.com/science/article/pii/S0921889022000343}
\BIBentrySTDinterwordspacing

\bibitem{JacobBeyond2020}
\BIBentryALTinterwordspacing
J.~Krantz, E.~Wijmans, A.~Majumdar, D.~Batra, and S.~Lee, ``Beyond the nav-graph: Vision-and-language navigation in continuous environments,'' \emph{CoRR}, vol. abs/2004.02857, 2020. [Online]. Available: \url{https://arxiv.org/abs/2004.02857}
\BIBentrySTDinterwordspacing

\bibitem{oquab2024dinov2learningrobustvisual}
\BIBentryALTinterwordspacing
S.~Liu, Z.~Zeng, T.~Ren, F.~Li, H.~Zhang, J.~Yang, C.~Li, J.~Yang, H.~Su, J.~Zhu, and L.~Zhang, ``Grounding {DINO}: Marrying {DINO} with grounded pre-training for open-set object detection,'' 2024. [Online]. Available: \url{https://openreview.net/forum?id=DS5qRs0tQz}
\BIBentrySTDinterwordspacing

\bibitem{GaoWangJingWangLi2024}
\BIBentryALTinterwordspacing
Y.~Gao, Z.~Wang, L.~Jing, D.~Wang, X.~Li, and B.~Zhao, ``\BIBforeignlanguage{en}{Aerial vision-and-language navigation via semantic-topo-metric representation guided llm reasoning},'' Oct. 2024, arXiv:2410.08500 [cs]. [Online]. Available: \url{http://arxiv.org/abs/2410.08500}
\BIBentrySTDinterwordspacing

\bibitem{gao2024embodied}
C.~Gao, B.~Zhao, W.~Zhang, J.~Zhang, J.~Mao, Z.~Zheng, F.~Man, J.~Fang, Z.~Zhou, J.~Cui, X.~Chen, and Y.~Li, ``Embodiedcity: A benchmark platform for embodied agent in real-world city environment,'' \emph{arXiv preprint}, 2024.

\bibitem{kamel2017model}
M.~Kamel, T.~Stastny, K.~Alexis, and R.~Siegwart, ``Model predictive control for trajectory tracking of unmanned aerial vehicles using robot operating system,'' \emph{Robot Operating System (ROS) The Complete Reference (Volume 2)}, pp. 3--39, 2017.

\bibitem{OPEN}
P.~Sopasakis and E.~Fresk, ``Optimization engine,'' \url{https://alphaville.github.io/optimization-engine/ }, 2024, accessed on December 4, 2024.

\bibitem{zhangMirage2022}
\BIBentryALTinterwordspacing
J.~Zhang, D.~Jin, and Y.~Li, ``Mirage: an efficient and extensible city simulation framework (systems paper),'' in \emph{Proceedings of the 30th International Conference on Advances in Geographic Information Systems}, ser. SIGSPATIAL '22.\hskip 1em plus 0.5em minus 0.4em\relax New York, NY, USA: Association for Computing Machinery, 2022. [Online]. Available: \url{https://doi.org/10.1145/3557915.3560950}
\BIBentrySTDinterwordspacing

\bibitem{shah2017airsimhighfidelityvisualphysical}
\BIBentryALTinterwordspacing
S.~Shah, D.~Dey, C.~Lovett, and A.~Kapoor, ``Airsim: High-fidelity visual and physical simulation for autonomous vehicles,'' 2017. [Online]. Available: \url{https://arxiv.org/abs/1705.05065}
\BIBentrySTDinterwordspacing

\bibitem{UE4}
A.~Sanders, \emph{An Introduction to Unreal Engine 4}.\hskip 1em plus 0.5em minus 0.4em\relax A K Peters/CRC Press, 2016.

\bibitem{anderson2018vision}
P.~Anderson, Q.~Wu, D.~Teney, J.~Bruce, M.~Johnson, N.~S{\"u}nderhauf, I.~Reid, S.~Gould, and A.~Van Den~Hengel, ``Vision-and-language navigation: Interpreting visually-grounded navigation instructions in real environments,'' in \emph{Proceedings of the IEEE conference on computer vision and pattern recognition}, 2018, pp. 3674--3683.

\bibitem{ku2020room}
A.~Ku, P.~Anderson, R.~Patel, E.~Ie, and J.~Baldridge, ``Room-across-room: Multilingual vision-and-language navigation with dense spatiotemporal grounding,'' \emph{arXiv preprint arXiv:2010.07954}, 2020.

\bibitem{FanChenJiangZhouZhangWang2023}
\BIBentryALTinterwordspacing
Y.~Fan, W.~Chen, T.~Jiang, C.~Zhou, Y.~Zhang, and X.~Wang, ``\BIBforeignlanguage{en}{Aerial vision-and-dialog navigation},'' in \emph{\BIBforeignlanguage{en}{Findings of the Association for Computational Linguistics: ACL 2023}}.\hskip 1em plus 0.5em minus 0.4em\relax Toronto, Canada: Association for Computational Linguistics, 2023, p. 3043–3061. [Online]. Available: \url{https://aclanthology.org/2023.findings-acl.190}
\BIBentrySTDinterwordspacing

\bibitem{8578485}
\BIBentryALTinterwordspacing
P.~Anderson, Q.~Wu, D.~Teney, J.~Bruce, M.~Johnson, N.~Sunderhauf, I.~Reid, S.~Gould, and A.~van~den Hengel, ``{ Vision-and-Language Navigation: Interpreting Visually-Grounded Navigation Instructions in Real Environments },'' in \emph{2018 IEEE/CVF Conference on Computer Vision and Pattern Recognition (CVPR)}.\hskip 1em plus 0.5em minus 0.4em\relax Los Alamitos, CA, USA: IEEE Computer Society, June 2018, pp. 3674--3683. [Online]. Available: \url{https://doi.ieeecomputersociety.org/10.1109/CVPR.2018.00387}
\BIBentrySTDinterwordspacing

\end{thebibliography}

\end{document}